\pdfoutput=1

\documentclass{article}

\PassOptionsToPackage{square,numbers}{natbib}
\usepackage[preprint]{neurips_2020}

\usepackage[utf8]{inputenc} % allow utf-8 input
\usepackage[T1]{fontenc}    % use 8-bit T1 fonts
\usepackage[bookmarks=false]{hyperref}       % hyperlinks
\usepackage{url}            % simple URL typesetting
\usepackage{booktabs}       % professional-quality tables
\usepackage{amsfonts}       % blackboard math symbols
\usepackage{nicefrac}       % compact symbols for 1/2, etc.
\usepackage{microtype}      % microtypography

\usepackage{amsmath}
\usepackage{float} 
\usepackage{graphicx}
\usepackage{caption} 

\title{Interpretable Visualizations with Differentiating Embedding Networks}

\author{%
  Isaac Robinson \\
  Yale University\\
  New Haven, CT 06510 \\
  \texttt{isaac.robinson@yale.edu} \\
}

\begin{document}

% \bibliography{resources}

\maketitle

% \begin{abstract}
%   We present a novel dimensionality reduction algorithm based on an unsupervised Siamese neural network training regime, called Siamese kernel networks (SKN). We use Siamese neural networks to create a kernel that embeds data in a space where the Euclidean distance between two samples corresponds to the probability their features are related. This embedding can be used for visualization or generalized dimensionality reduction. We also present a novel clustering algorithm that runs on top of the embedding, that without prior knowledge of the number of clusters is competitive with the state of the art clustering algorithms, all of which require prior knowledge of the number of clusters. Finally, we leverage this clustering algorithm to train an end-to-end neural network to predict cluster labels from the samples themselves, and then use SHAP scores to explain on a per-sample basis why that sample belongs to that cluster in terms of its features. This approach provides insight not offered by standard non-parametric data exploration techniques such as t-SNE or UMAP. We establish a new state of the art for clustering on FashionMNIST. Code will be released on GitHub shortly.
% \end{abstract}

\begin{abstract}
    We present a visualization algorithm based on a novel unsupervised Siamese neural network training regime and loss function, called Differentiating Embedding Networks (DEN). The Siamese neural network finds differentiating or similar features between specific pairs of samples in a dataset, and uses these features to embed the dataset in a lower dimensional space where it can be visualized. Unlike existing visualization algorithms such as UMAP or $t$-SNE, DEN is parametric, meaning it can be interpreted by techniques such as SHAP. To interpret DEN, we create an end-to-end parametric clustering algorithm on top of the visualization, and then leverage SHAP scores to determine which features in the sample space are important for understanding the structures shown in the visualization based on the clusters found. We compare DEN visualizations with existing techniques on a variety of datasets, including image and scRNA-seq data. We then show that our clustering algorithm performs similarly to the state of the art despite not having prior knowledge of the number of clusters, and sets a new state of the art on FashionMNIST. Finally, we demonstrate finding differentiating features of a dataset. Code available at \href{https://github.com/isaacrob/DEN}{this https url}.
\end{abstract}

\section{Introduction}

Exploratory data analysis is a primary application of unsupervised machine learning. Exploratory techniques such as clustering and visualization are indispensable when the underlying structure of data is unknown. With ever increasing volumes of complex data generated by, for example, biomedical research or edge devices, there is an increasing need for more sensitive data exploration tools.

Visualization techniques such as $t$-SNE \cite{maaten2008visualizing} and UMAP \cite{2018arXivUMAP} have gained wide usage as standard methods for exploratory data analysis. They attempt to approximate the manifold upon which data lies in two or three dimensions for visualization so that the structure of the data can be easily understood by researchers. Insights offered by such techniques are relied upon enough that they are used to help established the standard 'ground truth' in scRNA-seq datasets \cite{shekhar2016comprehensive} against which other approaches are judged. Furthermore, they have easily understood hyperparameters that allow people who are not experts in machine learning to successfully make use of them for whatever application they have in mind \cite{2018arXivUMAP, maaten2008visualizing}.

Clustering, another indispensable tool of exploratory data analysis, attempts to elucidate the deeper structure of the data by splitting it into discrete subsets linked by some characteristic. Visualization techniques are often used to gain insight into the clustered underlying structure of data, and indeed $t$-SNE can be shown to preserve clusters in its embedding \cite{linderman2019clustering}. However, they do not explicitly provide cluster labels for the data \cite{2018arXivUMAP, maaten2008visualizing}, so to successfully employ $t$-SNE or UMAP to generate clusters, a separate clustering algorithm has to be run on top of the $t$-SNE or UMAP visualization. This decouples the two exploratory data analysis techniques so that the embedding generated for visualization may not optimally reveal structures relevant for different clustering techniques. Nevertheless, clustering approaches such as Louvain tend to agree enough about the structure of the underlying data that both $t$-SNE and Louvain \cite{blondel2008fast} can be used to represent the same standard 'ground truth' \cite{shekhar2016comprehensive}.

While these approaches have offered invaluable insights into the structure of complex datasets, it remains a non-trivial task to understand which features of the data are important for understanding the revealed structures. While powerful model interpretation techniques exist \cite{ribeiro2016modelagnostic, NIPS2017_7062}, for the most part they depend on parametric representations of the models. $t$-SNE and UMAP work by moving representative points in a lower dimensional embedding according to a loss function \cite{2018arXivUMAP, maaten2008visualizing}, ultimately decoupling the visualization from interpretable features in the sample space, as they do not create a parametric relationship between their embedding and the samples they represent. Similarly, most clustering algorithms are non-parametric and do not create explicit, interpretable models for the data they attempt to represent. 

While there have been advances in parametric clustering models, mostly built around deep neural networks, most require prior knowledge of the number of clusters \cite{shaham2018spectralnet, mcconville2019n2d, mrabah2019deep, yang2016joint, mukherjee2019clustergan}, which severely limits their applicability to exploratory data analysis, or they rely on an intermediary non-parametric dimensionality reduction technique such as $t$-SNE \cite{ren2020deep}, which eliminates their parametric modeling interpretability.

\begin{figure}[]
    \centering
    \includegraphics[width=\textwidth]{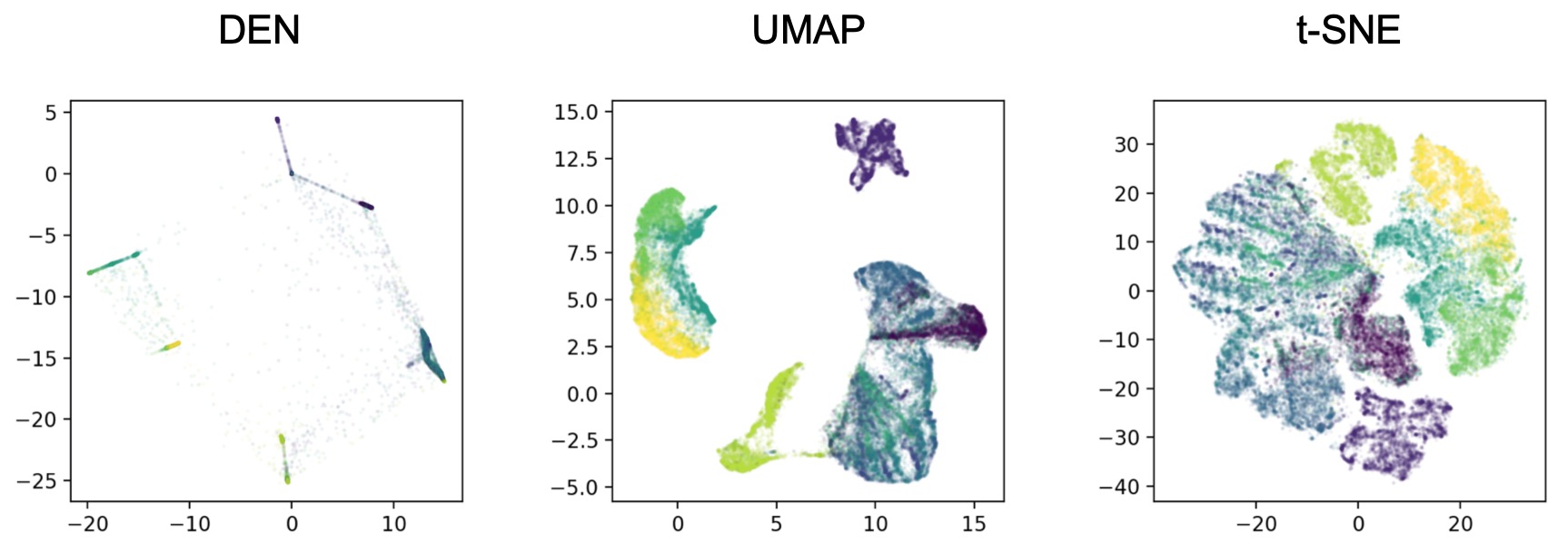}
    \caption{DEN embedding of FashionMNIST compared to UMAP and $t$-SNE embeddings.}
    \label{fig:fashion_mnist_embedding}
\end{figure}

In this work we introduce Differentiating Embedding Networks (DEN), an interpretable visualization and clustering algorithm that does not require prior knowledge of the number of clusters. Based on the success of SpectralNet for clustering, which uses a Siamese neural network to learn a kernel \cite{shaham2018spectralnet}, and $t$-SNE, which attempts to move points in a lower dimensional embedding such that the probability distribution between the lower dimensional points matches that of the data in sample-space \cite{maaten2008visualizing}, we create a Siamese neural network that learns a metric in the sample space and attempts to embed the data in a lower dimensional space according to a probability distribution to fit the learned metric. We learn the Siamese neural network metric as we embed the data, creating a parametric visualization tool. We then cluster the data and build a parametric model to take points from the embedding and predict which cluster they belong to, which, when combined with the Siamese network, creates an end-to-end parametric clustering algorithm. Finally, we use SHAP scores to determine which features in sample space are important \cite{NIPS2017_7062} for understanding the structures presented in the visualization as captured by the clustering algorithm. Our experiments show that DEN creates visualizations very similar to $t$-SNE and UMAP, but with tighter clusters. An example visualization compared to UMAP and $t$-SNE applied to the FasionMNIST dataset \cite{xiao2017fashion} can be seen in Figure \ref{fig:fashion_mnist_embedding}. We also show that our clustering results based off the DEN embedding perform similarly to the state of the art despite not having prior knowledge of the number of clusters. Finally, we demonstrate automatic determination of important and differentiating features for a dataset.

% Exploration of unlabeled data, including clustering and visualization, has become an essential part of biomedical research. In scRNA-seq research, for example, visualization with $t$-SNE and clustering with Louvain is often used to provide a standard labeling for cell types \cite{shekhar2016comprehensive}. Without prior knowledge of labels, it can be extremely difficult to understand the structure of the data at hand. 

\section{Methods}

In this section we describe the implementation and training regime for Siamese kernel networks. We describe the creation of an initial unweighted graph connecting various samples from the dataset motivated by the $t$-SNE adjacency matrix commonly used to accelerate $t$-SNE \cite{linderman2019fast}. We then describe the Siamese neural network and our novel loss function based on the $F$-distribution. Finally, we describe our method for interpreting the DEN visualization via clustering and SHAP scores.

\subsection{Building the Graph}

Like $t$-SNE, UMAP, and related techniques, we assume samples that are very close to each other according to some relevant metric in the sample space should be close to each other in the embedding space \cite{maaten2008visualizing, 2018arXivUMAP}. Similarly, we assume samples that are very far from each other in the sample space likely should be represented as far apart in the embedding space. Beyond that, we don't assume a metric such as the Euclidean distance in the sample space is well suited to describe a manifold upon which the data lies. To create a graph that leverages these assumptions and generate a training dataset for the Siamese kernel network, we label pairs of points as positive or negative, depending on if their distances in the sample space indicate that they should be close to or far from each other in the embedding space.

To determine pairs of positive points, we generate a $k$-nearest neighbors (KNN) graph based on some metric. We then create a shared nearest neighbor graph \cite{ertoz2002new} among each point's top $k$-nearest neighbors. In order to make sure each point is well represented, if a point does not have any shared nearest neighbors, we add its closest neighbors to the graph until it has at least $j$ neighbors, $j<k$. Therefore, each sample has at least $j$ positive samples associated with it. Both $j$ and $k$ are hyperparameters. Empirically, $j=1$ and $k=10$ work well.

To create negative pairs, for each sample we randomly create $k$ negative pairs where the probability of being paired with a particular other sample is proportional to the distance between the two samples according to the same metric. This approach is similar to the $D^2$ weighting used in \cite{arthur2006k}. We don't prune this graph at all. Note that, while $t$-SNE can be interpreted as the balancing of a repulsive and attractive force \cite{maaten2008visualizing}, where the repulsive force is universally applied to all pairs of samples, this graph deliberately selects only particular pairs to have a repulsive force between them.

% Note that the distance used to create pairs doesn't have to be the classic Euclidean distance; just like UMAP allows specification of different metrics \cite{2018arXivUMAP}, so can different metrics be used to generate the graph for Siamese kernel networks.

\subsection{Siamese Neural Network}

% For the purposes of embedding images and comparison with other approaches, we create a Siamese convolutional neural network (CNN) with four convolutions, followed by two fully connected layers. We use max pooling, ReLU activation, and BatchNorm \cite{ioffe2015batch}. However, this network does not have to be a CNN; to embed Cytof and scRNA-seq data, we use a fully connected network. The purpose is just to have a deep network that can create expressive features.

We use different neural network architectures to embed different types of data. In this way, we can ensure we are taking sufficient advantage of the neural network's ability to determine interesting features about the data, highlighting its distinction from existing manifold embedding techniques. To embed images, we use a standard convolutional neural network. To embed text and other token-based data, we use an embedding layer, take the average of the token vectors, and pass this vector through a hidden layer, much like fastText \cite{joulin2016bag}. For other kinds of data, including scRNA-seq, we use a standard fully connected deep neural network.

In creating our embedding space, we would ideally want the Euclidean distances between pairs of points to reflect how likely they are to be related based on the features learned by the Siamese neural network. We can interpret the last layer of a deep neural network with no following activation function as linear regression on the deep features learned by that network. Since the Euclidean distance between two points $x$ and $y$ in an $n$-dimensional embedding space is $||x - y||_2 = \sqrt{\sum_{i=1}^n (x_i-y_i)^2}$, we can interpret the square of the Euclidean distance as the sum of the squared residuals between $x$ and $y$. Since we're interpreting the final layer of the neural network as linear regression, we can argue that, if the learned features should ideally map $x$ and $y$ to the same part of the embedding, indicating a feature-based or semantic similarity between $x$ and $y$, then the residuals of the linear regression model should be approximately distributed according to an $F$-distribution \cite{ramsey1969tests}. 

To translate this interpretation into a loss function, we run a regression $F$-test and evaluate the cumulative density function (CDF) of the $F$-distribution with $d_1 = 1$ and $d_2 = n$ for the residual between $x$ and $y$. This parameter setting can be interpreted as the goodness of fit of each comparison between two $n$-dimensional embeddings being measured by the fit of $n$ one-dimensional regressions, which allows the network to cleanly take advantage of an increase in the dimension of the embedding space. From this, we get that the probability that $x$ and $y$ should be close to each other given their Euclidean distance $||x-y||_2 = d$ is 
    \[ P(x, y) = I_{\frac{d^2}{d^2 + n}}\bigg(\frac{1}{2}, \frac{n}{2}\bigg) \]
where $I_x(a, b)$ refers to the regularized incomplete beta function, defined as 
    \[ I_x(a, b) = \frac{B(x; a, b)}{B(a, b)}  \]
where $B(x; a, b)$ is the incomplete beta function and $B(a, b)$ is the beta function.

In order to quickly differentiate the incomplete beta function $B(x; a, b)$ as required to train a neural network with gradient descent, we let 
    \[ B(x; a, b) =\ _2F_1(a, 1-b, a+1, x) \frac{x^a}{a} \]  
and we then use a Laplace approximation of the hypergeometric function $_2F_1$ to make our loss function differentiable and closed-form \cite{butler2002laplace}.

Once we have $P(x, y)$, if $x$ and $y$ form a positive pair, we minimize $P$, and if they form a negative pair, we minimize $1-P$. Note that since $P$ is monotonic with the $F$-statistic, which is equal to the square of the Euclidean distance $||x-y||_2$, that this minimization is equivalent to minimizing the distance between positive pairs and maximizing the distance between negative pairs. This yields the desired property of the embedding space. It is worth noting that the $t$-distribution has been successful in $t$-SNE \cite{maaten2008visualizing}, and the square of $t$-distributed random variable is distributed according to an $F$-distribution.

\subsection{Interpretation}

We describe our method for interpreting DEN embeddings and visualizations. First, we cluster the embedding with spectral clustering \cite{von2007tutorial}. Then, we build a parametric model to predict cluster labels from the embedding. Finally, we turn the Siamese network and this clustering model into an end-to-end parametric model to assign cluster labels, and use SHAP scores to interpret which data features are important for placing a sample into a specific cluster \cite{NIPS2017_7062}.

\subsubsection{Clustering}

In order to facilitate exploratory data analysis, we want to create cluster labels on the embedding without prior knowledge of the number of clusters. Currently, most of the top performing clustering methods on a number of datasets require prior knowledge of the number of clusters \cite{mcconville2019n2d, mrabah2019deep, shaham2018spectralnet, mukherjee2019clustergan}, with a notable exception being DDC \cite{ren2020deep}. To address this issue, we use spectral clustering to cluster samples in the embedding space, which does not necessarily require prior knowledge of the number of clusters \cite{von2007tutorial}.

For our spectral clustering implementation, we first construct an affinity matrix using a Gaussian kernel in the embedding space from a subset of the total points. To speed up the spectral clustering algorithm, we only perform clustering on a subset of the data. The feasibility of performing spectral clustering only on a subset of the data is supported by the success of the Nystr{\"o}m method for kernel approximation \cite{williams2001using}. From the graph construction, we already have designated points that should be near each other in the embedding space, namely the positive pairs. Our affinity matrix is supposed to represent relatedness of points, so we can calculate the expected distance between positive pairs in the embedding space and use this as the bandwidth of our Gaussian kernel. This way, our affinity matrix captures the same definition of 'relatedness' we used to build our training samples. We calculate this expected distance by approximating the mean of the distance between all positive pairs in the embedding space. Calling this distance $d_{\text{avg}}$, our kernel can then be expressed as \[ k(x, y) = \text{exp}\bigg(-\frac{||x-y||_2^2}{\gamma d_{\text{avg}}^2}\bigg) \] where $x$ and $y$ are two points and $\gamma$ is a scaling factor. In practice, we let $\gamma = 1$, except when the clusters in the data are particularly small, as in the case of embedding the AG\_NEWS dataset in Figure \ref{fig:labeled_vis_comparisons}.

Calling this affinity matrix $A$, we compute the row sums and create a diagonal matrix with them called $D$. Using these matrices, we calculate the unnormalized graph Laplacian $L=D-A$ \cite{von2007tutorial}. To perform our spectral clustering, we compute the eigenvalues and eigenvectors of $L$, get an estimate for the number of clusters $k$ by counting how many eigenvalues are below a certain threshold, and then run the $k$-means algorithm on the first $k$ eigenvectors. We don't expect our graph components to be completely disconnected, so our threshold must be greater than $0$ \cite{von2007tutorial}. In practice, $10^{-2}$ seems to work well. This gives an overestimate of the number of clusters present since we impose such a weak threshold to count the number of clusters.

To deal with this overestimate and extend our cluster labels to samples outside of our subset used in the construction of the affinity matrix, we apply a KNN filter to the data. We train a KNN classifier to classify samples from the embedding space based on their cluster labels, with a relatively large number of neighbors. We then classify all points in the embedding with this classifier. The large number of neighbors works as a low-pass filter to remove high-frequency label signals, effectively denoising the labels domain. Labels that shift quickly among nearest neighbors are less likely to survive a KNN classifier, and are also more likely to be noise. This is similar to denoising via a low-passed graph Fourier transform \cite{6808520}. To the best of our knowledge, this is the first time a KNN classifier has been used to clean up spectral clustering labels in this way.

\subsubsection{Interpretation via Parametric Clustering}

Like the approaches used in N2D \cite{mcconville2019n2d} and DDC \cite{ren2020deep}, this spectral approach to clustering the DEN embedding is non-parametric and does not immediately support out-of-sample extension. Our goal is to create an interpretable visualization method, where the entire pipeline is parametric and applicable to new data, so we have to move beyond these non-parametric clustering approaches.

To address this problem, we create a second neural network to predict the cluster labels discovered by our spectral clustering approach from the embedding. Because we don't want to make any assumptions about the distribution of data in the embedding space, we use a self-normalizing neural network with the SeLU activation function \cite{klambauer2017self}. We train this network for 50 epochs. We then combine this clustering network with the Siamese network that generates the embedding in an end-to-end way, and fine-tune the whole network by training it for an additional 50 epochs at a lower learning rate. In this way, we create an end-to-end network that can take samples and predict which cluster they should belong to. To the best of our knowledge, this is the first fully parametric clustering approach that does not require prior knowledge of the number of clusters. As we now have a fully parametric clustering model, we can use SHAP scores to explain which features of a given sample are important for its being placed in a particular cluster \cite{NIPS2017_7062}, thus revealing which features are important for structures shown in the DEN visualization.

\section{Results}

Differentiating Embedding Networks place no limit on the dimension of the embedding space. Our experiments have shown no significant difference in clustering performance with different numbers of dimensions in the embedding space, but in theory, a higher dimensional representation could encapsulate more information about the structure of the data than a lower dimensional one. Having said that, for the sake of direct comparison with $t$-SNE and UMAP, which are frequently used to visualize data in two dimensions \cite{maaten2008visualizing, 2018arXivUMAP}, we present all of our results based on a two-dimensional embedding. First we present results of data visualizations, then of clustering, and finally of automatically-determined differentiating features of clustering results. Since algorithms for visualization cannot be directly compared with clustering algorithms, we compare against different algorithms in each section.

\subsection{Visualization}

% \begin{figure}[htp]
%     \centering
%     % \captionsetup{width=.9\linewidth}
%     \includegraphics[width=\textwidth]{figures/fashion_mnist_comparison_zoomed.jpg}
%     \includegraphics[width=\textwidth]{figures/usps_comparison.jpg}
%     \includegraphics[width=\textwidth]{figures/shekhar_comparison_zoomed.jpg}
%     \caption{Comparison of visualization techniques on different datasets.}
%     \label{fig:vis_comparisons}
% \end{figure}

\begin{figure}[]
    \centering
    \includegraphics[width=\textwidth]{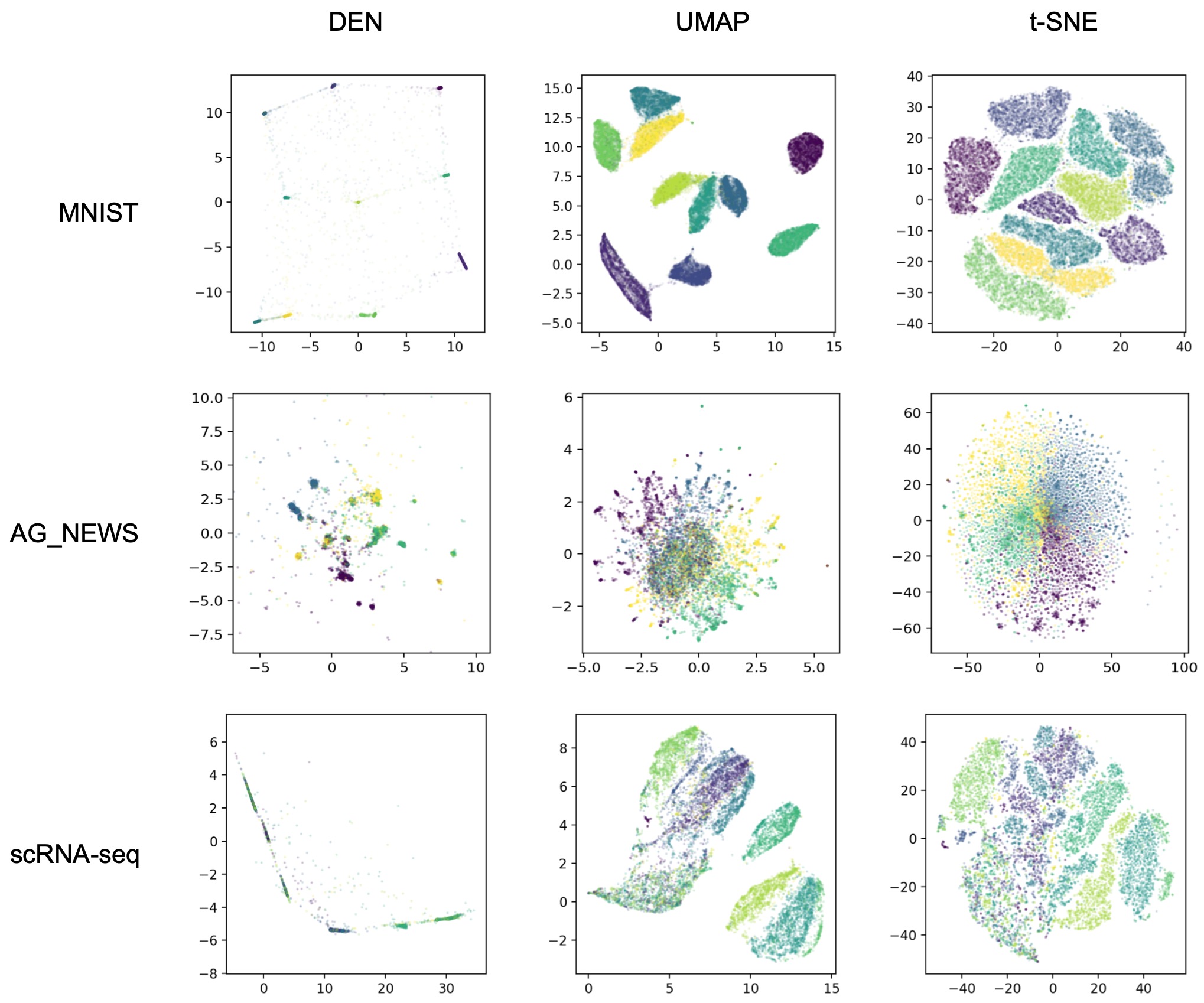}
    \caption{Comparison of visualization techniques on different datasets. Note DEN has been zoomed to the region of interest, leaving out a few outliers. MNIST is from \cite{lecun1998gradient}, AG\_NEWS is from \cite{gulli}, and the scRNA-seq dataset is from \cite{shekhar2016comprehensive}. We generated all figures ourselves.}
    \label{fig:labeled_vis_comparisons}
\end{figure}

\begin{figure}[]
    \centering
    \includegraphics[width=\textwidth]{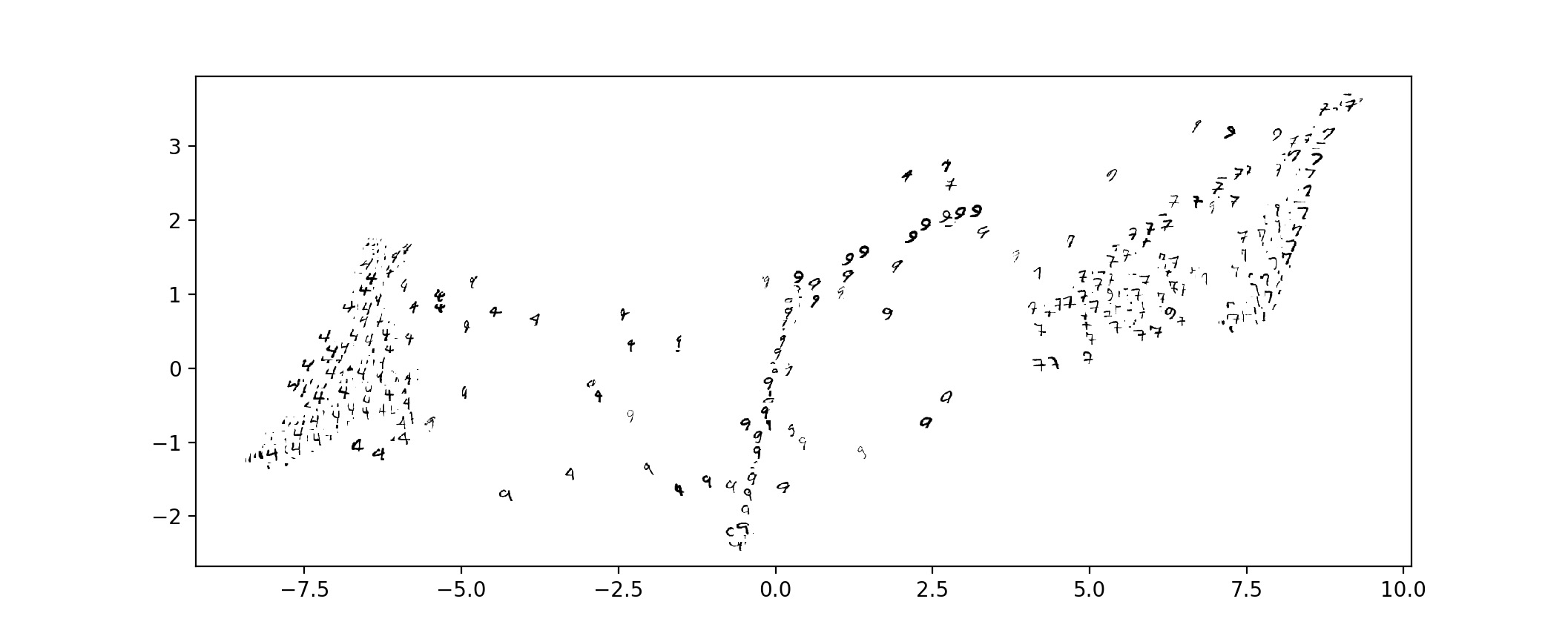}
    \caption{A DEN embedding of the digits 4, 7, and 9 from the MNIST dataset, with the images shown in their embedding locations. Note that we can see paths between clusters showing intermediate forms of digits.}
    \label{fig:mnist_embedding}
\end{figure}

Figure \ref{fig:labeled_vis_comparisons} shows visualizations produced by DEN, UMAP, and $t$-SNE, on three different datasets. Our approach tends to produce tighter clusters than either $t$-SNE or UMAP, with paths of samples bridging the gaps between the different clusters. When visualized as in Figure \ref{fig:mnist_embedding}, it becomes clear that these paths represent intermediate forms between the clusters based on the deep features extracted by the neural network. Since UMAP and $t$-SNE don't extract features, and instead attempt to represent the data manifold \cite{maaten2008visualizing, 2018arXivUMAP}, such paths won't show up in their visualizations. It should also be noted that our approach is scale-sensitive, in that running on a subset of the data produces a different representation for that data than would running it in the context of more data, because the selection of negative pairs is likely different. This scale sensativity is not present in $t$-SNE, and is much less pronounced in UMAP \cite{maaten2008visualizing, 2018arXivUMAP}. The difference in the presentations of the clusters containing digits 4, 7, and 9 in MNIST between Figures \ref{fig:labeled_vis_comparisons} and \ref{fig:mnist_embedding} show this scale sensativity; in Figure \ref{fig:mnist_embedding}, the clusters are much less concentrated than their counterparts in Figure \ref{fig:labeled_vis_comparisons}, where 4, 7, and 9 are found along the bottom of the MNIST visualization.

It should be noted that, in order to generate the MNIST and AG\_NEWS visualizations, we use feature extraction techniques that are unique to neural networks, notably convolutional neural networks and the architecture from fastText \cite{joulin2016bag}, respectively. These feature extractors allow us to build very different embeddings from those captured by UMAP and $t$-SNE, which operate in pixel space for MNIST and on TF-IDF vectors for AG\_NEWS. This highlights the potential of DEN for processing more complex data than is feasible with approaches that can't leverage the flexibility of neural networks. 

Each of the clusters in the AG\_NEWS visualization refers to a separate news subject, including differentiations between American and international sports news, between news pertaining to the Israeli/Palestinian conflict and civil unrest in a country, and between news pertaining to takeovers of tech companies and the interaction of tech companies and governments. These clusters are more specific than the labels provided with the dataset, which divide the data into World, Sports, Business, and Tech/Science.

\subsection{Clustering}

We compare the performance of our clustering algorithm against the current state of the art. Note that DEN, DDC-DA \cite{ren2020deep}, and UMAP+HDBSCAN are the only methods in the table that don't require prior knowledge of the number of clusters. When possible, we report both the accuracy (ACC) and normalized mutual information (NMI) of the algorithm, but if the number of clusters found is incorrect then we cannot report accuracy. Our algorithm comes in second of those in the table on USPS \cite{hull1994database} and MNIST \cite{lecun1998gradient}, and sets a new state of the art in terms of NMI on FashionMNIST, which is considered a substantially more difficult dataset \cite{xiao2017fashion}. These clustering results were produced based off a two-dimensional embedding that can also be used to visualize the data. With the exception of UMAP, no other algorithm in the table attempts to produce a visualization output. It should be noted that both N2D and DDC-DA do produce intermediate embeddings via UMAP or $t$-SNE that can be visualized, but the authors don't propose using their methods as generalized visualization approaches.

\begin{table}[htp]
  \caption{Clustering Performance (ACC/NMI)}
  \label{tab:cluster_results}
  \centering
  \begin{tabular}{lllllll}
    \toprule
    % \multicolumn{2}{c}{Part}                   \\
    % \cmidrule(r){1-2}
         & USPS & & MNIST & & FashionMNIST & \\
    \midrule
    DynAE & \textbf{0.981} & \textbf{0.948}  & \textbf{0.989} & \textbf{0.962} & 0.591 & 0.642   \\
    DDC-DA & 0.977 & 0.939 & 0.969 & 0.941 & 0.609 & 0.661 \\
    N2D     & 0.958 & 0.901 & 0.979 &  0.942  & \textbf{0.672} & \textit{0.684}    \\
    JULE     & 0.950 & 0.913 & 0.964 & 0.913 & - & - \\
    ClusterGAN     & 0.970 & 0.931 & 0.964  &  0.921 & - & - \\
    SpectralNet & - & - &  0.971 & 0.924 & - & - \\
    GDL & - & - & 0.964 & 0.910 & 0.627 & 0.660 \\
    UMAP+HDBSCAN & - & 0.877 & - & 0.884 & - & 0.594 \\
    \midrule
    Ours & \textit{0.979} & \textit{0.944} & \textit{0.984} & \textit{0.956} & \textit{0.635} & \textbf{0.710} \\
    \bottomrule
  \end{tabular}
%   \caption{Performances of various state of the art clustering methods on different datasets. * did not find the corrent number of clusters so did not compute accuracy.}
\end{table}

DynAE is from \cite{mrabah2019deep}, N2D is from \cite{mcconville2019n2d}, DDC-DA is from \cite{ren2020deep}, JULE is from \cite{yang2016joint}, ClusterGAN is from \cite{mukherjee2019clustergan}, SpectralNet is from \cite{shaham2018spectralnet}, GDL is from \cite{zhang2012graph}, and HDBSCAN is from \cite{mcinnes2017hdbscan}. Even though it isn't published as a standalone method, we put UMAP+HDBSCAN in our table because it is often discussed as a strong clustering method, and since UMAP is typically used as a visualization method, we felt it deserved direct comparison in terms of clustering performance. UMAP with HDBSCAN did not find the correct number of clusters on any dataset, so we did not calculate its accuracy, and in terms of its normalized mutual information score it did not outperform DEN.

All algorithms in the table except for GDL and UMAP are deep learning based. However, the only one that employs a similar neural network to ours is SpectralNet \cite{shaham2018spectralnet}, which trains a Siamese neural network \cite{koch2015siamese} to work as its kernel for its deep spectral clustering.

\subsection{Explanations}

\begin{figure}[htp]
    \centering
    \includegraphics[width=\textwidth]{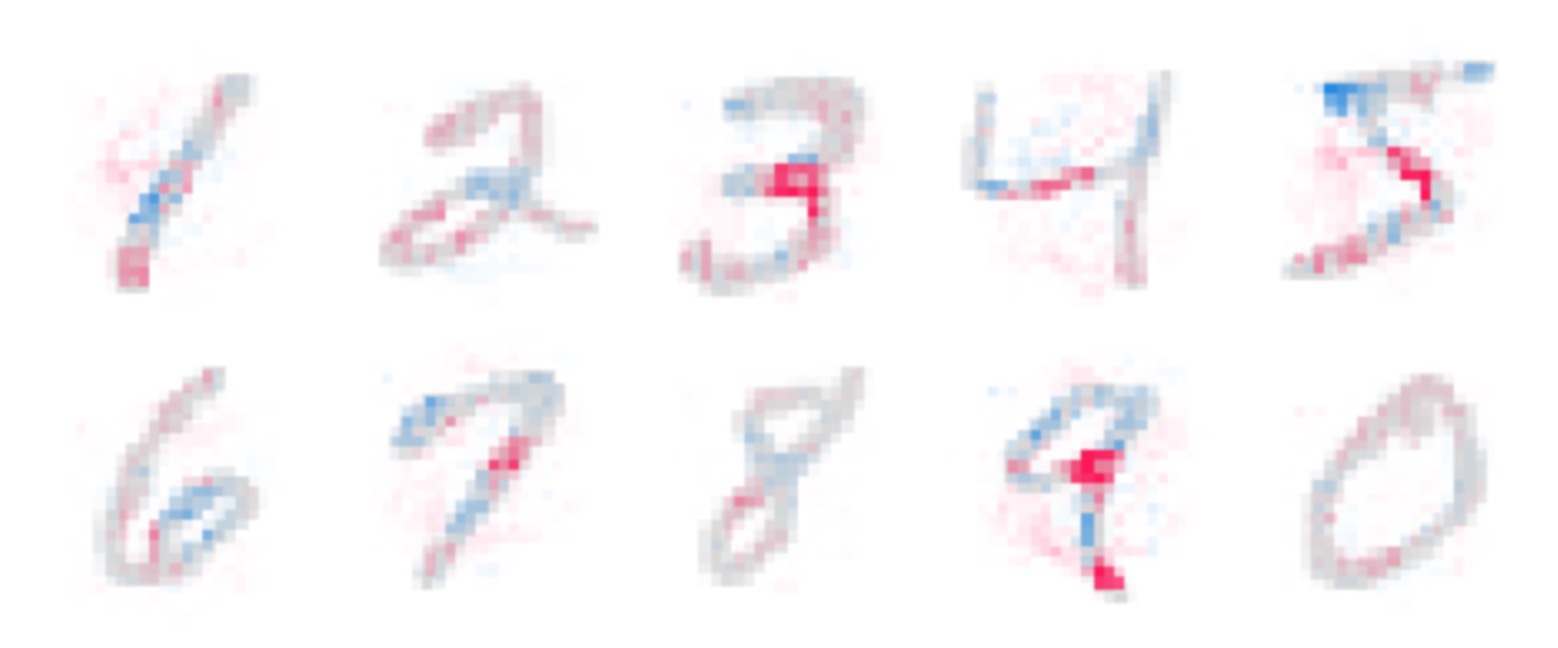}
    \caption{Explanations of what factors contribute to MNIST samples being placed into their appropriate clusters discovered by DEN, as determined by SHAP's DeepExplainer.}
    \label{fig:explanations}
\end{figure}

The fact that DEN determines on its own the optimal number of clusters and builds an end-to-end parametric model that can predict cluster labels means we can leverage SHAP to explain what characteristics of a sample contribute to its being placed in a particular cluster \cite{NIPS2017_7062}. This amounts to determining the defining characteristics of each sample that differentiate it from the rest of the dataset. Figure \ref{fig:explanations} shows such cluster label explanations applied to randomly selected examples of each MNIST digit. Red marks positive evidence for belonging to the cluster and blue marks negative evidence. Note that these descriptions are sample-specific. For example, the 1 is tilted to the side and might therefore be considered a 4. But the red coloring indicates that the missing arm from the 4 contributes strongly to its being clustered with the 1s instead. For the 5, we see three red blotches on the body corresponding to typical structures found in 5s. Furthermore, we see red blotches in curves where, if there had been curves present, it would have been classified as an 8, so the absence of those curves is significant for its classification as a 5. It is not visible in this display, but the dark blue on the upper left side of this particular 5 image that was chosen for analysis hides a jut that is atypical of 5s and therefore contributes evidence that perhaps this sample should not be a 5. Each of the other digits also has distinctive characteristics captured by DEN.

While displaying defining characteristics of images clearly demonstrates the underlying idea, the method is not limited to images. The same approach can be applied to determine which proteins or particular RNA expressions contribute to biological data being grouped in a particular way. Because $t$-SNE, UMAP, and other related methods are nonparametric \cite{maaten2008visualizing, 2018arXivUMAP}, the same approach cannot be applied to explain their cluster predictions \cite{NIPS2017_7062}. Conversely, the existing state of the art parametric clustering algorithms require prior knowledge of the number of clusters, and so can't be used for purely exploratory data analysis. In light of these facts, DEN occupies a potentially useful, unfilled niche.

\section{Conclusions}

Our novel approach to visualization, Differentiating Embedding Networks, shows promise for exploratory data visualization and clustering. It shows similar structure to that captured by UMAP and $t$-SNE, but with tighter clusters and the possibility of visualizing feature-based intermediate forms between clusters. For clustering, it performs similar to the current state of the art despite lacking prior knowledge of the number of clusters. We also set a new state of the art on FashionMNIST. Finally, DEN performs its clustering in an end-to-end parametric model, meaning we can explain why samples belong to particular clusters via SHAP scores. This allows us to visualize, cluster, and then identify defining characteristics of the data.

Future work includes creating a hierarchical organization approach taking advantage of DEN's inherent scale sensitivity, and creating a variant on DEN that can leverage labels to create an optionally semi-supervised algorithm, just as UMAP is able to leverage partially labeled data \cite{2018arXivUMAP}.

\clearpage

\pagebreak

\subsubsection*{Acknowledgments}

We'd like to thank Ariel Jaffe for his mentorship, encouragement, and advice; Yuval Kluger for his support and his offering of access to computing resources; and Stefan Steinerberger for essential inspiration and encouragement. Thank you.


\begin{thebibliography}{30}
\providecommand{\natexlab}[1]{#1}
\providecommand{\url}[1]{\texttt{#1}}
\expandafter\ifx\csname urlstyle\endcsname\relax
  \providecommand{\doi}[1]{doi: #1}\else
  \providecommand{\doi}{doi: \begingroup \urlstyle{rm}\Url}\fi

\bibitem[Arthur and Vassilvitskii(2006)]{arthur2006k}
D.~Arthur and S.~Vassilvitskii.
\newblock k-means++: The advantages of careful seeding.
\newblock Technical report, Stanford, 2006.

\bibitem[Blondel et~al.(2008)Blondel, Guillaume, Lambiotte, and
  Lefebvre]{blondel2008fast}
V.~D. Blondel, J.-L. Guillaume, R.~Lambiotte, and E.~Lefebvre.
\newblock Fast unfolding of communities in large networks.
\newblock \emph{Journal of statistical mechanics: theory and experiment},
  2008\penalty0 (10):\penalty0 P10008, 2008.

\bibitem[Butler et~al.(2002)Butler, Wood, et~al.]{butler2002laplace}
R.~W. Butler, A.~T. Wood, et~al.
\newblock Laplace approximations for hypergeometric functions with matrix
  argument.
\newblock \emph{The Annals of Statistics}, 30\penalty0 (4):\penalty0
  1155--1177, 2002.

\bibitem[Ertoz et~al.(2002)Ertoz, Steinbach, and Kumar]{ertoz2002new}
L.~Ertoz, M.~Steinbach, and V.~Kumar.
\newblock A new shared nearest neighbor clustering algorithm and its
  applications.
\newblock In \emph{Workshop on clustering high dimensional data and its
  applications at 2nd SIAM international conference on data mining}, pages
  105--115, 2002.

\bibitem[Gulli()]{gulli}
A.~Gulli.
\newblock Ag's corpus of news articles.
\newblock URL
  \url{http://groups.di.unipi.it/~gulli/AG_corpus_of_news_articles.html}.

\bibitem[Hull(1994)]{hull1994database}
J.~J. Hull.
\newblock A database for handwritten text recognition research.
\newblock \emph{IEEE Transactions on pattern analysis and machine
  intelligence}, 16\penalty0 (5):\penalty0 550--554, 1994.

\bibitem[Joulin et~al.(2016)Joulin, Grave, Bojanowski, and
  Mikolov]{joulin2016bag}
A.~Joulin, E.~Grave, P.~Bojanowski, and T.~Mikolov.
\newblock Bag of tricks for efficient text classification.
\newblock \emph{arXiv preprint arXiv:1607.01759}, 2016.

\bibitem[Klambauer et~al.(2017)Klambauer, Unterthiner, Mayr, and
  Hochreiter]{klambauer2017self}
G.~Klambauer, T.~Unterthiner, A.~Mayr, and S.~Hochreiter.
\newblock Self-normalizing neural networks.
\newblock In \emph{Advances in neural information processing systems}, pages
  971--980, 2017.

\bibitem[Koch et~al.(2015)Koch, Zemel, and Salakhutdinov]{koch2015siamese}
G.~Koch, R.~Zemel, and R.~Salakhutdinov.
\newblock Siamese neural networks for one-shot image recognition.
\newblock In \emph{ICML deep learning workshop}, volume~2. Lille, 2015.

\bibitem[LeCun et~al.(1998)LeCun, Bottou, Bengio, and
  Haffner]{lecun1998gradient}
Y.~LeCun, L.~Bottou, Y.~Bengio, and P.~Haffner.
\newblock Gradient-based learning applied to document recognition.
\newblock \emph{Proceedings of the IEEE}, 86\penalty0 (11):\penalty0
  2278--2324, 1998.

\bibitem[Linderman and Steinerberger(2019)]{linderman2019clustering}
G.~C. Linderman and S.~Steinerberger.
\newblock Clustering with t-sne, provably.
\newblock \emph{SIAM Journal on Mathematics of Data Science}, 1\penalty0
  (2):\penalty0 313--332, 2019.

\bibitem[Linderman et~al.(2019)Linderman, Rachh, Hoskins, Steinerberger, and
  Kluger]{linderman2019fast}
G.~C. Linderman, M.~Rachh, J.~G. Hoskins, S.~Steinerberger, and Y.~Kluger.
\newblock Fast interpolation-based t-sne for improved visualization of
  single-cell rna-seq data.
\newblock \emph{Nature methods}, 16\penalty0 (3):\penalty0 243--245, 2019.

\bibitem[Lundberg and Lee(2017)]{NIPS2017_7062}
S.~M. Lundberg and S.-I. Lee.
\newblock A unified approach to interpreting model predictions.
\newblock In I.~Guyon, U.~V. Luxburg, S.~Bengio, H.~Wallach, R.~Fergus,
  S.~Vishwanathan, and R.~Garnett, editors, \emph{Advances in Neural
  Information Processing Systems 30}, pages 4765--4774. Curran Associates,
  Inc., 2017.
\newblock URL
  \url{http://papers.nips.cc/paper/7062-a-unified-approach-to-interpreting-model-predictions.pdf}.

\bibitem[Maaten and Hinton(2008)]{maaten2008visualizing}
L.~v.~d. Maaten and G.~Hinton.
\newblock Visualizing data using t-sne.
\newblock \emph{Journal of machine learning research}, 9\penalty0
  (Nov):\penalty0 2579--2605, 2008.

\bibitem[McConville et~al.(2019)McConville, Santos-Rodriguez, Piechocki, and
  Craddock]{mcconville2019n2d}
R.~McConville, R.~Santos-Rodriguez, R.~J. Piechocki, and I.~Craddock.
\newblock N2d:(not too) deep clustering via clustering the local manifold of an
  autoencoded embedding.
\newblock \emph{arXiv preprint arXiv:1908.05968}, 2019.

\bibitem[McInnes et~al.(2017)McInnes, Healy, and Astels]{mcinnes2017hdbscan}
L.~McInnes, J.~Healy, and S.~Astels.
\newblock hdbscan: Hierarchical density based clustering.
\newblock \emph{Journal of Open Source Software}, 2\penalty0 (11):\penalty0
  205, 2017.

\bibitem[{McInnes} et~al.(2018){McInnes}, {Healy}, and
  {Melville}]{2018arXivUMAP}
L.~{McInnes}, J.~{Healy}, and J.~{Melville}.
\newblock {UMAP: Uniform Manifold Approximation and Projection for Dimension
  Reduction}.
\newblock \emph{ArXiv e-prints}, Feb. 2018.

\bibitem[Mrabah et~al.(2019)Mrabah, Khan, Ksantini, and
  Lachiri]{mrabah2019deep}
N.~Mrabah, N.~M. Khan, R.~Ksantini, and Z.~Lachiri.
\newblock Deep clustering with a dynamic autoencoder: From reconstruction
  towards centroids construction.
\newblock \emph{arXiv preprint arXiv:1901.07752}, 2019.

\bibitem[Mukherjee et~al.(2019)Mukherjee, Asnani, Lin, and
  Kannan]{mukherjee2019clustergan}
S.~Mukherjee, H.~Asnani, E.~Lin, and S.~Kannan.
\newblock Clustergan: Latent space clustering in generative adversarial
  networks.
\newblock In \emph{Proceedings of the AAAI Conference on Artificial
  Intelligence}, volume~33, pages 4610--4617, 2019.

\bibitem[Ramsey(1969)]{ramsey1969tests}
J.~B. Ramsey.
\newblock Tests for specification errors in classical linear least-squares
  regression analysis.
\newblock \emph{Journal of the Royal Statistical Society: Series B
  (Methodological)}, 31\penalty0 (2):\penalty0 350--371, 1969.

\bibitem[Ren et~al.(2020)Ren, Wang, Li, and Xu]{ren2020deep}
Y.~Ren, N.~Wang, M.~Li, and Z.~Xu.
\newblock Deep density-based image clustering.
\newblock \emph{Knowledge-Based Systems}, page 105841, 2020.

\bibitem[Ribeiro et~al.(2016)Ribeiro, Singh, and
  Guestrin]{ribeiro2016modelagnostic}
M.~T. Ribeiro, S.~Singh, and C.~Guestrin.
\newblock Model-agnostic interpretability of machine learning, 2016.

\bibitem[{Sandryhaila} and {Moura}(2014)]{6808520}
A.~{Sandryhaila} and J.~M.~F. {Moura}.
\newblock Discrete signal processing on graphs: Frequency analysis.
\newblock \emph{IEEE Transactions on Signal Processing}, 62\penalty0
  (12):\penalty0 3042--3054, 2014.

\bibitem[Shaham et~al.(2018)Shaham, Stanton, Li, Nadler, Basri, and
  Kluger]{shaham2018spectralnet}
U.~Shaham, K.~Stanton, H.~Li, B.~Nadler, R.~Basri, and Y.~Kluger.
\newblock Spectralnet: Spectral clustering using deep neural networks.
\newblock \emph{arXiv preprint arXiv:1801.01587}, 2018.

\bibitem[Shekhar et~al.(2016)Shekhar, Lapan, Whitney, Tran, Macosko, Kowalczyk,
  Adiconis, Levin, Nemesh, Goldman, et~al.]{shekhar2016comprehensive}
K.~Shekhar, S.~W. Lapan, I.~E. Whitney, N.~M. Tran, E.~Z. Macosko,
  M.~Kowalczyk, X.~Adiconis, J.~Z. Levin, J.~Nemesh, M.~Goldman, et~al.
\newblock Comprehensive classification of retinal bipolar neurons by
  single-cell transcriptomics.
\newblock \emph{Cell}, 166\penalty0 (5):\penalty0 1308--1323, 2016.

\bibitem[Von~Luxburg(2007)]{von2007tutorial}
U.~Von~Luxburg.
\newblock A tutorial on spectral clustering.
\newblock \emph{Statistics and computing}, 17\penalty0 (4):\penalty0 395--416,
  2007.

\bibitem[Williams and Seeger(2001)]{williams2001using}
C.~K. Williams and M.~Seeger.
\newblock Using the nystr{\"o}m method to speed up kernel machines.
\newblock In \emph{Advances in neural information processing systems}, pages
  682--688, 2001.

\bibitem[Xiao et~al.(2017)Xiao, Rasul, and Vollgraf]{xiao2017fashion}
H.~Xiao, K.~Rasul, and R.~Vollgraf.
\newblock Fashion-mnist: a novel image dataset for benchmarking machine
  learning algorithms.
\newblock \emph{arXiv preprint arXiv:1708.07747}, 2017.

\bibitem[Yang et~al.(2016)Yang, Parikh, and Batra]{yang2016joint}
J.~Yang, D.~Parikh, and D.~Batra.
\newblock Joint unsupervised learning of deep representations and image
  clusters.
\newblock In \emph{Proceedings of the IEEE Conference on Computer Vision and
  Pattern Recognition}, pages 5147--5156, 2016.

\bibitem[Zhang et~al.(2012)Zhang, Wang, Zhao, and Tang]{zhang2012graph}
W.~Zhang, X.~Wang, D.~Zhao, and X.~Tang.
\newblock Graph degree linkage: Agglomerative clustering on a directed graph.
\newblock In \emph{European Conference on Computer Vision}, pages 428--441.
  Springer, 2012.

\end{thebibliography}
\end{document}